\pdfoutput=1

\documentclass[11pt]{article}

\usepackage{EMNLP2023}

\usepackage{times}
\usepackage{latexsym}
\usepackage{amssymb}

\usepackage[T1]{fontenc}

\usepackage[utf8]{inputenc}

\usepackage{microtype}

\usepackage{graphicx}       
\usepackage{multirow}       
\usepackage{booktabs}       
\usepackage{enumitem}       
\usepackage{subcaption}     
\usepackage[hang,flushmargin]{footmisc}  
\usepackage{inconsolata}
\usepackage{hyperref}
\usepackage{multirow}
\hypersetup{hidelinks}
\hypersetup{colorlinks=true}
\usepackage{array}
\usepackage{bm}

\newcommand{\LN}{\linebreak\noindent}
\newcommand{\MBE}{\texttt{MBE}}
\newcommand{\SBM}{\texttt{SBM}}
\newcommand{\LSG}{\texttt{LSG}}
\newcommand{\MSG}{\texttt{MSG}}
\newcommand{\DBM}{\texttt{DBM}}

\setenumerate[1]{leftmargin=*}    
\setitemize[1]{leftmargin=*}      

\title{What is Your Favorite Gender, MLM?\\ Gender Bias Evaluation in Multilingual Masked Language Models}

\author{Jeongrok Yu, Seong Ug Kim, Jacob Choi, Jinho D. Choi \\
  Department of Computer Science \\
  Emory University, Atlanta, GA \\
  \texttt{\{jyu248,seong.kim,jcho535,jinho.choi\}@emory.edu}\\
  }

\begin{document}
\maketitle

\begin{abstract}

Bias is a disproportionate prejudice in favor of one side against another. 
Due to the success of transformer-based Masked Language Models (MLMs) and their impact on many NLP tasks, a systematic evaluation of bias in these models is needed more than ever.
While many studies have evaluated gender bias in English MLMs, only a few works have been conducted for the task in other languages.
This paper proposes a multilingual approach to estimate gender bias in MLMs from 5 languages: Chinese, English, German, Portuguese, and Spanish.
Unlike previous work, our approach does not depend on parallel corpora coupled with English to detect gender bias in other languages using multilingual lexicons.
Moreover, a novel model-based method is presented to generate sentence pairs for a more robust analysis of gender bias, compared to the traditional lexicon-based method.
For each language, both the lexicon-based and model-based methods are applied to create two datasets respectively, which are used to evaluate gender bias in an MLM specifically trained for that language using one existing and 3 new scoring metrics.
Our results show that the previous approach is data-sensitive and not stable as it does not remove contextual dependencies irrelevant to gender.
In fact, the results often flip when different scoring metrics are used on the same dataset, suggesting that gender bias should be studied on a large dataset using multiple evaluation metrics for best practice.
\end{abstract}
\section{Introduction}
\label{sec:introduction}

The advent of transformer models \cite{vaswani} and subsequent development of contextualized embedding encoders \cite{devlin-etal-2019-bert,DBLP:journals/corr/abs-1907-11692} have led to the widespread deployment of large language models for crucial societal tasks \cite{hartvigsen2022toxigen}.
However, the use of such models has brought to light critical concerns regarding bias \cite{bender-friedman-2018-data,dixon:18a,hutchinson-etal-2020-social}.
Despite efforts to enhance Masked Language Models (MLMs) \cite{Clark2020ELECTRA:, tan-bansal-2019-lxmert}, adapted to pre-train the transformers for language understanding by predicting masked tokens from their context, the use of sophisticated models and extensive datasets has intensified worries about bias in MLMs.

The ubiquity of language models in society has sparked a growing interest in detecting and mitigating their inherent biases.
Detecting gender disparities in language technologies has gained traction across multiple domains, as evidenced by studies on identifying human-like biases in a transformer encoder \cite{kurita-etal-2019-measuring} and creating benchmarks for testing gender bias \cite{zhao-etal-2018-gender}.
These works have recognized undesirable biases from language models, leading to more studies on revealing bias in embeddings \cite{blodgett-etal-2020-language}.
Indeed, language models can have bias issues that must be addressed to ensure equitable and inclusive outcomes \cite{bender, sun-etal-2019-mitigating}.

There has been pioneering work on evaluating gender bias in static word embeddings using word analogies \cite{bolukbasi2016man}, while the evaluation of gender bias in contextualized word embeddings has largely focused on monolingual models.
\citet{zhao-etal-2018-learning} analyzed word pairs to differentiate which words contain gender information/bias.
\citet{liang-etal-2020-monolingual} examined the MLMs' tendency to predict gendered pronouns associated with certain occupations.
There have also been innovative debiasing methods for MLMs, such as re-balancing the corpus by switching bias attribute words \cite{unknown} or using derivation to normalize sentence vectors from MLMs \cite{bommasani-etal-2020-interpreting}. 
Nonetheless, there remains a gap in research on gender bias in multilingual MLMs, which has not been explored to the same extent.
This presents a crucial area for research to ensure that multilingual language models are free from harmful biases.

This paper improves the gender bias evaluation in multilingual MLMs by addressing the limitations of previous work.
Section~\ref{sec:approach} identifies such limitations and presents an enhanced method to generate sentence pairs for gender bias evaluation in MLMs.
Section~\ref{sec:making the most} provides multilingual lexicons to detect sentences with gendered words for four languages.
Section~\ref{sec:experiments} compares the performance of our method to ones from existing methods.
Finally, Section~\ref{sec:analysis} shows that our method retains more data from the target corpus, and is more consistent than the other methods in evaluating gender bias, especially when the data is skewed in the gender distribution.
Our main contributions are as follows:\footnote{All our resources, including datasets and evaluation scripts, are publicly available through our open source project: \url{https://github.com/anonymous}.}

\begin{itemize}

\item We create multilingual gender lexicons to detect sentences with gendered words in Chinese, English, German, Portuguese, and Spanish without relying on parallel datasets, which enables us to extract more diverse sets of gendered sentences and facilitate more robust evaluations.

\item We present two novel metrics and methods that provide a rigorous approach to comparing multilingual MLMs and datasets. Our approach ensures meaningful and fair comparisons, leading to more reliable and comprehensive assessments of gender bias in multilingual MLMs.

\end{itemize}
\section{Related Work}
\label{sec:related-work}

The issue of bias in large language models has garnered significant attention in recent years, prompting several studies that aim to assess and address the presence of bias in MLMs.
\citet{nangia-etal-2020-crows} introduced CrowS-Pairs, a dataset consisting of single sentences with masked attribute words, aiming to assess potential social bias in terms of race, gender, and religion in MLMs.
\citet{nadeem-etal-2021-stereoset} adopted a similar approach by masking modified tokens to measure bias.
These studies were confined\LN to English, however, and lacked a precise definition of the term ``bias'', resulting in ambiguity and assumptions about its meaning \cite{blodgett-etal-2021-stereotyping}.


\citet{ahn-oh-2021-mitigating} introduced a novel approach to evaluating bias in MLMs using pairs of sentences with varying degrees of masking. 
They proposed a new metric, the Categorical Bias score, showing the variance of log-likelihood interpreted as an effect size of the attribute word. 
In addition, they analyzed ethnic bias across six languages, making an effort to generalize bias evaluation.
However, their method still required human-written sentences with bias annotations, which is a limitation in terms of capturing the natural usage of language and can be exploited when the model finds a simple loophole around the set of rules \cite{durmus-etal-2022-spurious}.
Our work requires minimal annotation and is evaluated on a real dataset rather than a contrived one.

A few studies have attempted to evaluate bias in multilingual MLMs. 
\citet{kaneko-AULA} assessed bias by utilizing a set of English words associated with males and females and computing their likelihoods.
\citet{kaneko-etal-2022-gender} used a parallel corpus in English and eight other languages, where bias was annotated solely in English, to evaluate gender bias in the target language models.
Our approach is distinguished in that it does not rely on a parallel corpus, making it language-independent.
\section{Methodology}
\label{sec:approach}

\begin{figure*}[ht]
\centering
\includegraphics[width=\textwidth]{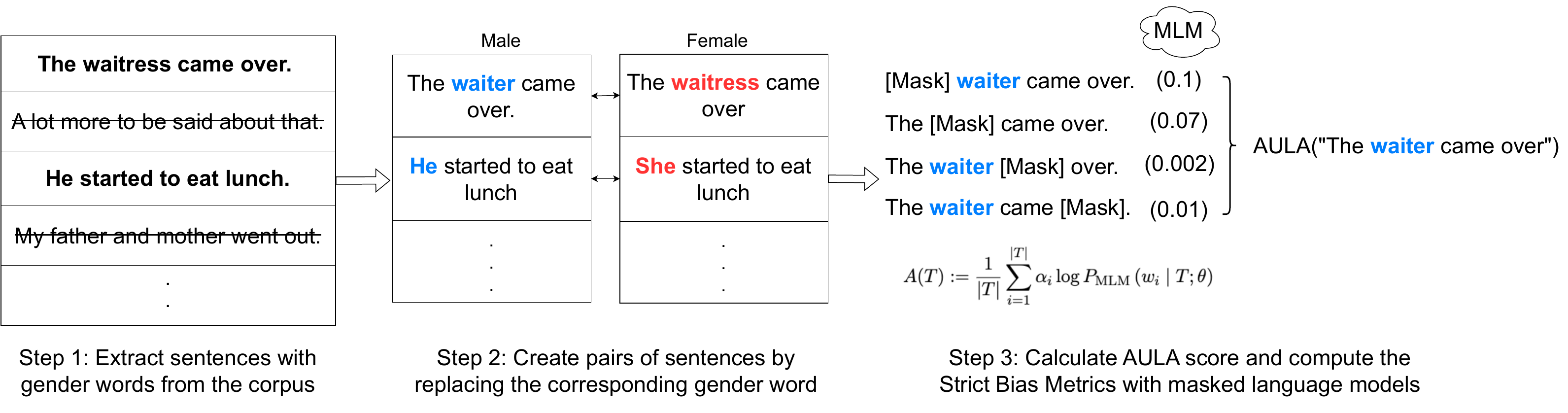}
\caption{An overview of the lexicon-based sentence extraction (Section~\ref{sec:rule}).}
\label{fig:lexicon-based}
\end{figure*}

\subsection{Multilingual Bias Evaluation}
\label{sec:baseline}

\citet{kaneko-etal-2022-gender} proposed a Multilingual Bias Evaluation (\MBE) score to assess the gender bias in MLMs across many languages using a multilingual corpus with English translations.
To ensure a fair comparison with previous work, we adopt the \MBE\ score as our baseline, which employs a three-step approach to detect potential gender bias.


First, \MBE\ scans English sentences containing male or female nouns.
These nouns are identified using a set of gendered nouns from \citet{bolukbasi2016man} and common first names from \citet{nangia-etal-2020-crows}.
\MBE\ extracts the corresponding sentence in the target language, under the premise that the parallel sentence also contains gendered terms.
The extracted sentences are categorized into \(\mathcal{T}_f\) and \(\mathcal{T}_m\), representing sets of female and male sentence.


For each sentence $S = [w_1, .., w_n] \in (\mathcal{T}_f \cup \mathcal{T}_m)$ where $w_i$ is the $i$'th token in $S$, \MBE\ estimates its All Unmasked Likelihood with Attention (AULA; \citet{kaneko-AULA}).
Given all tokens in $S$, except for $w_i$, and pre-trained parameters $\theta$, the likelihood of an MLM predicting $w_i$ is measured by $P_{\mathrm{MLM}}\left(w_i \mid S\backslash w_i ; \theta\right)$.
AULA is then computed by summing the log-likelihoods of all $w_i$ multiplied by\LN the averages of the multi-head attention weights $\alpha_i$ associated with $w_i$ as in Equation~\ref{AULA}:\vspace{-0.1em}
\begin{equation}\label{AULA}
A(S) = \frac{1}{|S|} \sum_{\forall {w_i \in S}} \alpha_i \cdot \log P_{\mathrm{MLM}}\left(w_i \mid S\backslash w_i ; \theta\right)
\end{equation}
Thus, AULA estimates the relative importance of individual words in the sentence by the MLM.

%

\noindent Finally, the \MBE\ score is measured by comparing all sentences between \(T_f\) and \(T_m\) as in Equation~\ref{MBE}:
\begin{equation}\label{MBE}
\frac{\displaystyle\sum_{\scriptscriptstyle{\forall S_m \in \mathcal{T}_m}} \sum_{\scriptscriptstyle{\forall S_f \in \mathcal{T}_f}} \gamma\left(S_m, S_f \right)\cdot \mathbb{I} \left(S_m, S_f\right)}{\displaystyle\sum_{\scriptscriptstyle{\forall S_m \in \mathcal{T}_m}} \sum_{\forall \scriptscriptstyle{S_f \in \mathcal{T}_f}} \gamma\left(S_m, S_f \right)}
\end{equation}
\(\gamma(S_f, S_m)\) is the cosine similarity score between sentence embeddings of \(S_f\) and \(S_m\).
The indicator function $\mathbb{I}(S_m, S_f)$ returns $1$ if $A\left(S_m \right)>A \left(S_f \right)$; otherwise, $0$.
Therefore, the \MBE\ score shows the percentage of male sentences that the MLM prefers over female sentences in the parallel corpus.


%


\subsection{Strict Bias Metric}
\label{sec:strict}

\noindent A significant drawback of \MBE\ is its potential lack of rigor in comparisons.
One key limitation lies in the use of AULA to calculate sentence likelihoods\LN that can lead to tokens unrelated to gendered words exerting undue influence on the score.
This is even more problematic when comparing AULAs for sentences that are notably dissimilar.


To address unexpected measurement errors in evaluating MLM bias, we propose the Strict Bias Metric (\SBM), which solely compares likelihoods between parallel sentences differ only by gendered words, minimizing potential confounding factors.
As the number of comparisons in AULAs decreases when using \SBM, it allows for a more focused and targeted analysis so that \SBM\ ensures a meaningful assessment by capturing the likelihood differences incurred only by variations in gender words.
The \SBM\ score is measured as in Equation~\ref{SBM}:
\begin{equation}\label{SBM}
\frac{\sum_{(S_m, S_f) \in (\mathcal{T}_m \times \mathcal{T}_f)}\mathbb{I}\left(S_m, S_f\right)}{|\mathcal{T}_m|}
\end{equation}
Since \SBM\ requires the identification of the specific gender words, we create a multilingual lexicon that provides sets of gendered words in five languages (Section~\ref{sec:Gender_lexicon}).
Moreover, since \SBM\ needs sentence pairs differ only by gender words, we present two methods, lexicon-based (Section~\ref{sec:rule}) and model-based (Section~\ref{sec:model-based}), to generate such sentence pairs using solely monolingual datasets.

%
%

\subsection{Lexicon-based Sentence Generation}
\label{sec:rule}

\noindent The lexicon-based sentence generation (\LSG) first identifies every sentence containing a single gender word and constructs its counterpart by replacing the gender word with its opposite gendered word provided in the lexicon (Section~\ref{sec:Gender_lexicon}).
For the sentence, ``The \textit{waitress} came over'', \LSG\ generates the new sentence ``The \textit{waiter} came over'' by replacing the gender word (Figure \ref{fig:lexicon-based}).
\SBM\ is then computed between these two sentences, taking into account the relative importance of those gender words.

Sentences containing multiple gender words are excluded in this approach because it is ambiguous to identify them as either male- or female-gendered sentences.
For example, a sentence ``The \textit{actor} fell in love with the \textit{queen}'' is considered neither a male nor female sentence because it contains words for both genders and masking one of them for an MLM would make this sentence to be the opposite gender.\LN
Including such sentences would undermine the reliability of the \SBM\ score.
Thus, sentences of this nature are excluded from our experiments to ensure the robustness of bias evaluation.

The corpus may have an imbalanced distribution of sentences with male and female words that can result in a bias in the overall data representation.
To address this issue, we extract an equal number of male and female sentences such that $|\mathcal{T}_m| = |\mathcal{T}_f|$.
It is worth mentioning that we tried to discard sentences exhibiting strong contextual biases towards a specific gender from our evaluation set (e.g., The \textit{princess} gave birth to twins), as they may introduce inherent biases and compromise the results.

%

\subsection{Model-based Sentence Generation}
\label{sec:model-based}

\begin{figure}[h]
\centering
\includegraphics[width=\columnwidth]{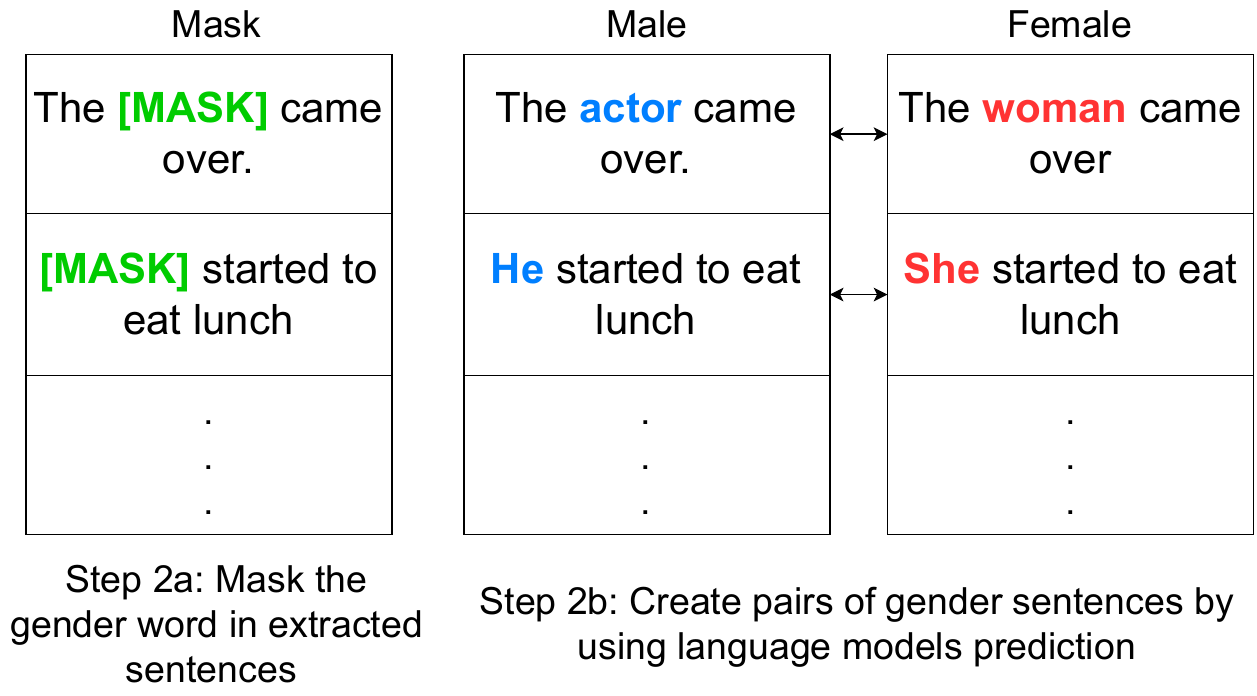}
\caption{An overview of the model-based sentence extraction (Section~\ref{sec:model-based}) using the examples in Figure~\ref{fig:lexicon-based}.}
\label{fig:model-based}
\end{figure}

\noindent Figure \ref{fig:model-based} describes an overview of our model-based sentence generation (\MSG).
Similar to \LSG, it begins by collecting sentences with only one gender word.
It then masks the gender word in each sentence and employs an MLM to predict the most likely male and female words based on the context (e.g., the first row in Figure~\ref{fig:model-based}).
This enables \MSG\ to select the most probable gender words for every sentence by MLM's predictions, ensuring that the comparisons are conducted on model-derived sentence pairs.


\noindent In theory, \MSG\ can always generate both male and female words for any masked word by taking words from the corresponding gender sets whose MLM scores are the highest among the others in the sets.
However, in practice, \MSG\ may fail to generate the sentence pair because the MLM does not predict ``meaningful'' gender words with high confidence.
In this case, \MSG\ adapts \LSG\ to handle the missing sentence.
For the second row example in Figure~\ref{fig:model-based}, the MLM effectively predicts the female word `\textit{She}' but does not predict any male word with high confidence, in which case, it takes the opposite gendered word `\textit{He}' as its lexicon-based counterpart.
On the other hand, sentences for which the MLM fails to predict both male or female words with high scores are discarded.
Once all sentence pairs are created, \SBM\ is applied to assess potential bias in the MLM towards any particular gender.

\begin{figure}[h]
\centering
\includegraphics[width=\columnwidth]{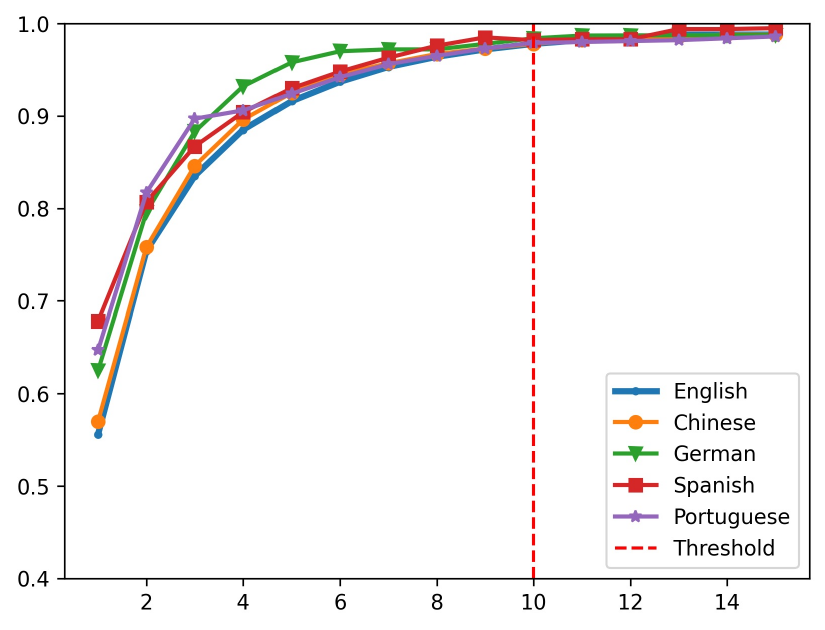}
\caption{The proportions of sentences containing gender words whose confidence scores are higher than the threshold $0.01$ covered by the $k = [1, 15]$ predictions.}
\label{fig:k=10}
\end{figure}

For our experiments, a threshold of $0.01$ is used to determine whether MLM's prediction confidence is high or not.
This threshold is observed by manual inspection of MLM predictions in five languages.
In addition, we use the threshold to configure the top-$k$ reliable predictions for MLMs.
To determine the optimal value for $k$, an iterative analysis is conducted with $k$ ranging from 1 to 15, assessing how many predictions above the threshold are covered by the top $k$ predictions.
Our findings indicate that $k = 10$ is the optimal choice, as the convergence rate of the model exhibits only marginal changes beyond this cutoff point (Figure~\ref{fig:k=10}).


\subsection{Direct Comparison Bias Metric}
\label{sec:direct_comparison}

\noindent \SBM\ quantifies the extent to which MLMs prefer to predict male words over female words given the neutral context.
Since MLMs provide confidence scores for the predicted words, it is also possible to make word-level comparisons.
Thus, we propose the Direct Comparison Bias Metric (\DBM) that compares the scores of the predicted gender words for every (predictable) sentence.

Given a sentence containing one gender word, an MLM is employed to predict male and female words.
If the scores for both male and female word predictions are below the threshold (Section~\ref{sec:model-based}), the sentence is excluded as the comparison, in this case, is not meaningful.
Then, \DBM\ is measured by replacing the indicator function $\mathbb{I}(S_m, S_f)$ in Eq.~\ref{SBM} with $\mathbb{I}_d(w_m, w_f)$, where $w_m \in S_m$ is the predicted male word and $w_f \in S_f$ is the predicted female word.
$\mathbb{I}_d$ returns $1$ if the MLM's prediction score of $w_m$ is higher than that of $w_f$; otherwise, $0$.
\section{Data Preparation}
\label{sec:making the most}

\subsection{Multilingual Gender Lexicon}
\label{sec:Gender_lexicon}

\noindent We leverage the male and female word list provided by \citet{bolukbasi2016man} to construct our English gender word set.
This set consists of pairs of words with each gender word having its counterpart of the opposite gender, which makes it suitable for \LSG\ (\textsection\ref{sec:rule}) and \MSG\ (\textsection\ref{sec:model-based}).
Unlike \citet{kaneko-etal-2022-gender}, we exclude common first names from the CrowS-Pairs dataset \cite{nangia-etal-2020-crows} when making our lexicon.
This is done to avoid potential issues related to transliteration during name translation, which can lead to MLMs treating the first names as phonetically similar words in the target language.


Our Multilingual Gender Lexicon (MGL) is compiled by translating gender words in the English set using automatic systems, such as Bing\footnote{Microsoft Bing Translator: \url{https://www.bing.com/translator}}, DeepL\footnote{DeepL Translate: \url{https://www.deepl.com/translator}}, and Google\footnote{Google Translate: \url{https://translate.google.com}}, into the following eight languages: Arabic, Chinese, German, Indonesian, Japanese, Portuguese, Russian, and Spanish.
The automatic translations are carefully evaluated by three native speakers of each language.
If the majority of the reviewers find a translation to be unnatural or gender-neutral, both the translation and its counterpart are\LN excluded from MGL.
For example, a pronoun `\textit{sie}' in German has two meanings: `she' and `you' (honorific).
Although `\textit{sie}' commonly refers to a female, it was removed from our lexicon to leave no room for ambiguity regarding its connotation of gender information.
This meticulous process ensures that MGL contains accurate and natural translations of only gendered words in each language.


\subsection{MGL Validation}
\label{sec:gl_val}

\noindent The coverage of MGL is assessed by comparing the words in parallel sentences extracted from the TED corpus (2020 v1; \citet{reimers-2020-multilingual-sentence-bert}).\footnote{TED2020 v1: \url{https://opus.nlpl.eu/TED2020.php}}
The TED corpus consists of about 4,000 TED talks that comprise a total of 427,436 English sentences.
Many of the English sentences have been translated into 100+ languages by certified translators.

For the validation, a set $\mathcal{E}_r$ of 11,000 English sentences are randomly sampled from the TED corpus.
The sentences in $\mathcal{E}_r$ are checked against the English gender word set (Section~\ref{sec:Gender_lexicon}) to create another set $\mathcal{E}_g$ of English gendered sentences.
Next, for every target language $\ell$, a set $\mathcal{G}^\ell$ is created by finding $\ell$'s translations of the sentences in $\mathcal{E}_g$ from the corpus.
Not all sentences in $\mathcal{E}_g$ may come with translations such that $|\mathcal{G}^\ell| \leq |\mathcal{E}_g|$.
Finally, a new set $\mathcal{G}^\ell_g$ of gendered sentences is created where each sentence in $\mathcal{G}^\ell_g$ includes the translated gender word according to the target-language gender word (Section~\ref{sec:Gender_lexicon}).
Table \ref{table:1} illustrates the results from this validation, including the coverage percentages of MGL for the eight target languages.



\noindent MGL's coverage rates in gender words are below $0.5$ for the following four languages: Indonesian, Russian, Japanese and Arabic.
Many gender words in English are translated into gender-neutral words in Indonesian because it would sound unnatural to use gender-specific translations \cite{dwiastuti-2019-english}.
Both Russian and Arabic are morphologically-rich languages with extensive inflectional/derivational systems \cite{al-haj-lavie-2010-impact, rozovskaya-roth-2019-grammar}, which contribute to their low coverage rates.
Japanese is a pro-drop langauge allowing the omission of the subject in a sentence, which makes identifying gender words challenging because the subject pronouns, such as `\textit{he}' and `\textit{she}', are often dropped in natural discourse.


\begin{table}[h]
\small\centering
\begin{tabular}{c|r|r|c}
\toprule
\bf Language & \multicolumn{1}{c|}{$\bm{|\mathcal{G}^\ell|}$} & \multicolumn{1}{c|}{$\bm{|\mathcal{G}^\ell_g|}$} & \bf Coverage (\%)\\
\midrule
German     & 1,226 & 1,124 & \bf 91.7 \\ 
Spanish    & 1,380 & 1,125 & \bf 81.5 \\
Portuguese & 1,206 &   928 & \bf 76.9 \\
Chinese    & 1,325 &   997 & \bf 75.2 \\
\midrule
Indonesian &   671 &   312 & 46.5 \\
Russian    & 1,289 &   583 & 45.2 \\
Japanese   & 1,288 &   466 & 36.6 \\
Arabic     & 1,327 &   252 & 19.0 \\
\bottomrule
\end{tabular}
\caption{The coverage rates of sentences containing gender words from MGL across the eight target languages.}
\label{table:1}
\end{table}

\noindent It is important to note that the assumption made by \citet{kaneko-etal-2022-gender} that the gender information in an English sentence containing a gender word is retained in parallel sentences in the target languages, even if the corresponding gender words do not exist in the parallel sentences, may not hold true for certain languages mentioned above (e.g., Indonesian, Japanese).
Thus, our bias evaluation uses only languages with high coverage rates: English, German, Spanish, Portuguese, and Chinese.


\subsection{Sentence Pair Generation}
\label{sec:dataset}

\noindent Two sets of sentence pairs are created to evaluate gender bias of MLMs, one using \LSG\ and the other using \MSG, in English as well as the top-4 languages in Table~\ref{table:1}, all of which contain sufficient numbers of gendered words available in MGL.
For our experiments, BERT-based language-specific transformer encoders are used for English \cite{devlin-etal-2019-bert}, German \cite{chan-etal-2020-germans}, Spanish \cite{CaneteCFP2020}, Portuguese \cite{Souza}, and Chinese \cite{cui-etal-2020-revisiting}.
The statistics of these datasets in comparison to the one used by \citet{kaneko-etal-2022-gender} are presented in Table~\ref{tab0}.


\begin{table*}[htp!]
\centering\small{ 
\begin{tabular}{c|cc|ccr}
\toprule
\bf Language & \textbf{Kaneko}\bm{$_{org}$} & \textbf{Kaneko}\bm{$_{all}$} & \textbf{\LSG} & \textbf{\MSG} & \multicolumn{1}{c}{\bf Total} \\
\midrule
English    &   --- & 39,040 & 25,993 & 28,112 &  34,970 \\
Chinese    & 6,800 & 36,270 & 22,196 & 22,616 &  30,547 \\
German     & 4,700 & 26,639 & 32,436 & 29,667 &  33,154 \\
Portuguese & 5,700 & 29,975 & 24,608 & 31,670 &  36,072 \\
Spanish    & 7,100 & 37,808 & 76,972 & 96,995 & 114,168 \\
\bottomrule
\end{tabular}}
\caption{The statistics of gender bias evaluation datasets. \textbf{Kaneko}\bm{$_{org}$}: \# of gendered sentences used by \citet{kaneko-etal-2022-gender}. \textbf{Kaneko}\bm{$_{all}$}: \# of all gendered sentences extracted from the TED corpus utilizing the same method as in \citet{kaneko-etal-2022-gender}. Note that the previous work used only a subset of \textbf{Kaneko}\bm{$_{all}$} for their evaluations. \textbf{\LSG\ \& \MSG}:\LN \# of sentence pairs generated by lexicon-based (\textsection\ref{sec:rule}) and model-based (\textsection\ref{sec:model-based}) methods, respectively; the numbers in these columns should be doubled for a fair comparison with the numbers in \textbf{Kaneko}\bm{$_*$}. \textbf{Total}: \# of all extracted sentences from the TED corpus by using MGL (before discarding any sentences for balancing).}
\label{tab0}
\end{table*}


\noindent All sentences consisting of single gender words are extracted from the TED corpus by leveraging MGL.
For \LSG, an equal number of male and female sentences are extracted for the creation of each dataset in Table~\ref{tab0} to minimize potential contextual bias that arises from imbalanced distributions between male and female sentences (Section~\ref{sec:rule}). 
However, this constrains the extracted sentences to the size of the smaller gender category.
For example, if there exist 100 male sentences but only 50 female sentences,\LN half of the male sentences are discarded to match the number of female sentences.
Consequently, the total number of extracted sentences becomes 100, not 150.
In this case, even if there is contextual bias present in those sentences, the balanced number of sentences ensures that it does not significantly affect the overall evaluation results.


\begin{table}[h]
\centering\small
\begin{tabular}{cccc}
\toprule
\bf Language & \bf Both& \bf One & \bf None \\
\midrule
English    & 63.8\% & 16.6\% & 19.6\% \\
Chinese    & 59.1\% & 14.9\% & 26.0\% \\
Spanish    & 51.9\% & 33.1\% & 15.0\% \\
Portuguese & 43.0\% & 44.8\% & 12.2\% \\
German     & 30.7\% & 58.8\% & 10.5\% \\
\bottomrule
\end{tabular}
\caption{The proportions of sentences extracted using MGL, for which \MSG\ generates sentences for \textbf{both} genders, \textbf{one} gender, and \textbf{none}.}
\label{tab3}
\end{table}

\noindent For \MSG, on the other hand, the MLM generates both male and female sentences for the majority of extracted sentences.
Even with input that the MLM generates only one gender sentences, the ratio between male and female sentences produced by \MSG\ is often more balanced than that of \LSG.
For the previous example where 150 sentences are extracted (100 male and 50 female), assume that the MLM generates both male and female sentences for 60, only male sentences for 40, only female sentences for 30, and no sentences for 20.
\MSG\ discards the 20 no-generated sentences and 10 (out of the 40) male-only sentences, so its count matches to the one for the female-only sentences, resulting in a retention of 120 sentences, where the ratio between male and\LN female sentences is even.
Table~\ref{tab3} shows the proportions among the extracted sentences \MSG\ generates for both genders, one gender, and none.


For languages such as German, Portuguese, and Spanish, it is important to ensure grammatical correctness in the sentences generated by \LSG\ and \MSG. 
It is because these languages have gendered articles, adjectives, demonstratives, possessives, and attributive pronouns that need to agree with the genders of the generated words. 
For German, 19.85\% of the extracted sentences contain such gendered components.
To address this, we add mappings between gender words and their dependent gendered components to MGL for those languages.
We then use the mappings to replace gendered components in the generated sentences accordingly. 
For example, in a German sentence ``ein guter \textit{Mann}'' (a good \textit{man}), when \textit{Mann} (\textit{man}) is replaced with \textit{Frau} (\textit{woman)}, the article \textit{ein} is also replaced with \textit{eine} using the mapping provided in MGL.
This ensures that the generated sentences maintain agreement between gender words and their associated components.

\section{Experiments}
\label{sec:experiments}

We utilize the TED parallel corpus (Section~\ref{sec:gl_val}) to build a comprehensive corpus for the evaluation of gender bias in the transformer-based multilingual language models (Section \ref{sec:dataset}).
For evaluation, the transformer implementation of \citet{wolf-etal-2020-transformers} is employed.
All experiments are conducted on an Apple M1-Pro Chip with a 14-core GPU. The entire evaluation process, which includes processing all sentence pairs in every language, is completed in about two hours with efficient performance.


\begin{table*}[!h]
\centering\small{ 
\begin{tabular}{c|cc|ccc|c} 
\toprule
\bf Language & \textbf{Kaneko}\bm{$_{org}$} & \textbf{Kaneko}\bm{$_{all}$} & \textbf{\LSG} & \textbf{\MSG} & \textbf{\DBM} & \bf Male / Female (Ratio) \\
\midrule
English    & ---   & 52.07 ($\pm$1.34) & 50.39 ($\pm$0.28) & 45.49 & 75.18 & 62.83 / 37.17 (1.69:1) \\
Chinese    & 52.86 & 46.67 ($\pm$0.55) & 46.42 ($\pm$0.68) & 53.15 & 89.62 & 63.67 / 36.33 (1.75:1) \\
German     & 54.69 & 45.78 ($\pm$1.72) & 52.31 ($\pm$0.64) & 55.43 & 44.72 & 48.92 / 51.08 (0.96:1) \\
Portuguese & 53.07 & 46.70 ($\pm$0.81) & 51.77 ($\pm$0.44) & 61.04 & 73.36 & 65.89 / 34.11 (1.93:1) \\
Spanish    & 51.44 & 48.52 ($\pm$1.04) & 41.68 ($\pm$0.76) & 50.74 & 72.34 & 66.29 / 33.71 (1.97:1) \\
\bottomrule
\end{tabular}}
\caption{Left Group: Multilingual Bias Evaluation (\MBE) scores evaluated on the sub-sampled sentences (\textbf{Kaneko}\bm{$_{org}$}) and all sentences in the TED corpus (\textbf{Kaneko}\bm{$_{all}$}; $\pm$: standard deviation). Middle Group: The Strict Bias Metric (\SBM) scores achieved by \LSG\ and \MSG, as well as the Direct Comparison Bias Metrics (\DBM) scores obtained by \MSG. Right Group: The distributions of (and the ratios between) male and female sentences in the TED corpus (in \%).}
\label{tab5}
\end{table*}

\subsection{Multilingual Bias Evaluation on Kaneko\bm{$_*$}}
\label{sec:gb_score}

\noindent Upon replicating \citet{kaneko-etal-2022-gender}, we observe that the previous study utilized only $\approx$25\% of the sentences in the TED corpus (Table~\ref{tab0}).
For a more reliable evaluation, we reconstruct these datasets by following their methods using all sentences in the corpus, and compute the \MBE\ scores (Section~\ref{sec:baseline}).
Table~\ref{tab5} shows the results from \citet{kaneko-etal-2022-gender} (Kaneko$_{org}$), as well as the results evaluated on our reconstructed datasets (Kaneko$_{all}$).\footnote{The evaluation on Kaneko$_{all}$ is conducted in five-folds, similar to \LSG, for the same reason as explained in Section~\ref{sec:rule_results}.}

Surprisingly, while the English MLM shows bias towards male terms in Kaneko$_{all}$, MLMs in all the other four languages exhibit bias towards female terms, which contradicts the findings in Kaneko$_{org}$.
The discrepancy between the results obtained from Kaneko$_{org}$ and Kaneko$_{all}$ highlights the limitations of working with smaller datasets. 


\subsection{Strict Bias Metric for \LSG\ and \MSG}
\label{sec:rule_results}

To assess gender bias using \LSG, for each language, we create five folds of evaluation datasets by randomly truncating the larger gender set (Section~\ref{sec:dataset}), while keeping the sentences in the smaller gender set unchanged across all folds.
Hence, sentences in the larger gender set vary across the different folds.
Table~\ref{tab5} presents the \SBM\ scores (Section~\ref{sec:strict}) of this evaluation for the five languages.
Our results reveal that English, German, and Portuguese MLMs are biased towards males, with the scores greater than 0.5.
In contrast, Chinese and Spanish MLMs depict a bias towards females.
These trends are consistent across all five evaluation folds; thus reinforcing the reliability of our findings.


For \MSG, the \SBM\ scores are also used to assess gender bias.
Unlike \LSG, only a small portion of the larger gender sets get truncated using this method; thus, we create a single fold for the \MSG\ evaluation.
Our results show that the English MLM has a bias towards females, although the other MLMs show a bias towards males.
Interestingly, the findings from \LSG\ and \MSG\ disagree in terms of the bias directions for English, Chinese, and Spanish. 
Several factors may attribute to this, such as differences in the pretrained MLMs or characteristics of the languages, which we will explore in the future.


\subsection{Direct Comparison Bias Metric for \MSG}
\label{direct_result}

\noindent Finally, we assess gender bias using the \DBM\ metric (Section~\ref{sec:direct_comparison}) on the \MSG\ datasets. 
The scores obtained by this metric are considerably more extreme\LN than those from \MBE\ and \SBM.
Notice that the \DBM\ scores align with the gender distributions shown in\LN the TED corpus (Table~\ref{tab5}), implying that using \DBM\ to evaluate gender bias of MLMs on a corpus with a significant gender imbalance can lead to unreliable\LN results, although they are suitable for quantifying the bias in the evaluation corpus itself.

\section{Analysis}
\label{sec:analysis}

\subsection{Performance Analysis}
\label{sec: performance}



\noindent The discrepancy in results between Kaneko$_{org}$ and Kaneko$_{all}$, as discussed in Section~\ref{sec:Gender_lexicon}, is attributed to the subsampling strategy, which is employed to balance the number of male and female sentences for a more robust evaluation.
Ironically, it resulted in a less stable evaluation because there is no guarantee that a randomly selected set of sentences from the larger gender set would exhibit similar bias as another randomly selected set from the same group.
This issue becomes more challenging when the size of the subsampled set is significantly smaller than the original gender set.
Thus, minimizing the size gap between the selected set and the original set is crucial for conducting a robust bias evaluation.


It is worth noting that the standard deviations of the \SBM\ scores from \LSG\ are noticeably smaller than those of Kaneko$_{all}$ in Table~\ref{tab5}, except for Chinese.\LN
This variance can be problematic when the score is\LN close to the threshold of 50, potentially resulting in a contradictory finding.
In this regard, \MSG\ has an advantage over \MBE\ and \LSG\ as it generally discards significantly fewer sentences compared to the other two methods (Table~\ref{tab10}) by generating new sentences for both genders, offering a more stable method for evaluating gender bias.


\begin{table}[htp!]
\centering\small
\begin{tabular}{c|ccc}
\toprule
\bf Lang & \textbf{\MBE} & \textbf{\LSG} & \textbf{\MSG} \\
\midrule
English    & 31.98 & 25.67 & \textbf{19.60} \\
Chinese    & 32.13 & 27.34 & \textbf{25.96} \\
German     & 35.35 & \textbf{2.17} & 10.51 \\
Portuguese & 30.90 & 31.78 & \textbf{12.20} \\
Spanish    & 31.99 & 32.58 & \textbf{15.04} \\
\bottomrule
\end{tabular}
\caption{The percentages of the discarded sentences for balancing both genders. \MSG\ discards a significantly smaller number of sentences compared to the other two methods, except for German where \LSG\ shows the least amount of truncation.}
\label{tab10}
\vspace{-0.5em}
\end{table}

\begin{figure}[htp!]
\centering
\includegraphics[width=\columnwidth]{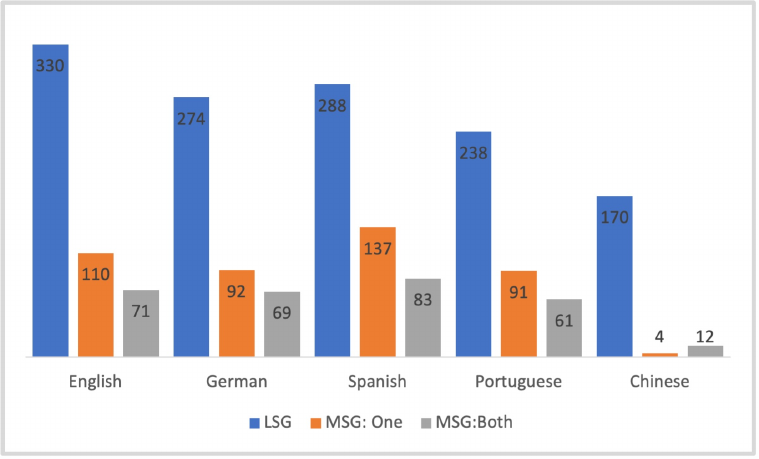}
\caption{The numbers of gender word types generated by \LSG\ and \MSG. \MSG: One|Both: the number of gender word types where \MSG\ generates for only one gender or both genders, respectively.}
\label{bargraph-crop}
\vspace{-0.5em}
\end{figure}

\noindent While \MSG\ provides a more consistent evaluation with a lower truncation rate compared to the other methods, it may end up producing less diversity in\LN gender words than \LSG.
Figure~\ref{bargraph-crop} shows that \LSG\ extracts sentences with a greater number of unique gender words across the five languages than \MSG.
The proportion of unique gender words used to fill the \texttt{[Mask]} does not exceed 50\% for all languages except for Chinese, where the MLM fills over 18,000 sentences using only 12 male and 4 female words.
By incorporating a diverse set of vocabulary from one-gender predictions obtained using \LSG, \MSG\ can quantify the bias in MLMs even in sentences they are not inclined to generate.


\subsection{Discussion}
\label{sec: discussion}

One major limitation of our evaluation framework is its restriction to the selected five languages.
To extend the applicability of \LSG\ \% \MSG\ to additional languages, further enhancements can be made to MGL by leveraging a verified parallel corpus and a word alignment tool.
While the proposed methods have effectively extract gendered sentences based on the lexicons, it is worth noting that gender can also be conveyed through other linguistic features.
This highlights the importance of conducting in-depth morphological analysis in languages where gender is grammatically encoded in articles, verbs, adjectives, etc.
Another limitation is that our evaluation of gender bias has been conducted only on language-specific models.
Exploring gender bias in different languages using cross-lingual language models could yield interesting findings compared to the language-specific models.


MGL opens up opportunities for future research to assess gender bias in a wide range of MLMs beyond those investigated in this study.
Moreover, evaluating gender bias in any corpus becomes feasible, as it is not limited to a parallel corpus.
Rather than examining MLM gender bias solely through translations from a parallel corpus, using a corpus in the language of interest can lead to more meaningful results.
This study also raises the question of whether the length of sentences in language models impacts the gender bias scores, suggesting further investigation for a conclusive answer.

\section{Conclusion}
\label{sec:conclusion}

\noindent This paper presents robust methods for evaluating gender bias in masked language models across five\LN languages: English, Chinese, German, Portuguese, and Spanish.
Using our multilingual gender lexicon (MGL), three evaluation metrics: multilingual bias evaluation (\MBE), strict bias metric (\SBM), and direct comparison bias metric (\DBM), and two sample generation methods: lexicon-based (\LSG) and model-based (\MSG), a comprehensive analysis is conducted, revealing that \MSG\ is the most generalizable and consistent method.

As bias evaluation is a rapidly evolving field with the emergence of new methods and metrics, we emphasize the importance of a collaborative effort from diverse perspectives to advance this research.
To establish an unbiased bias evaluation system, it is essential to approach it from multiple angles.
We hope that our work contributes to ongoing endeavors aimed at addressing gender bias and serves as an inspiration for further exploration in this critical area of research.

%

\bibliography{custom}
\bibliographystyle{acl_natbib}

\end{document}